\pdfoutput=1
\documentclass[11pt]{article}

\usepackage{acl}

\usepackage{times}
\usepackage{latexsym}

\usepackage[T1]{fontenc}
\usepackage[utf8]{inputenc}

\usepackage{microtype}
\usepackage{booktabs}
\usepackage{graphicx}
\usepackage{multirow}
\usepackage{ulem}
\title{ESimCSE: Enhanced Sample Building Method for Contrastive Learning of Unsupervised Sentence Embedding}

\author{
    Xing Wu\textsuperscript{\rm 1,2,3}, Chaochen Gao\textsuperscript{\rm 1,2}\thanks{The first two authors contribute equally.}, Liangjun Zang\textsuperscript{\rm 1},  Jizhong Han\textsuperscript{\rm 1},  Zhongyuan Wang\textsuperscript{\rm 3}, Songlin Hu\textsuperscript{\rm 1,2}\thanks{Corresponding author.}
    \\
    \textsuperscript{\rm 1}Institute of Information Engineering, Chinese Academy of Sciences, Beijing, China\\
    \textsuperscript{\rm 2}School of Cyber Security, University of Chinese Academy of Sciences, Beijing, China\\
    \textsuperscript{\rm 3}Kuaishou Technology, Beijing, China
    \\
    \{gaochaochen,zangliangjun,hanjizhong,husonglin\}@iie.ac.cn
    \\\{wuxing,wangzhongyuan\}@kuaishou.com
}

\begin{document}
\maketitle
\begin{abstract}
SimCSE \footnote{We focus on unsupervised sentence embedding, so SimCSE in this article refers to unsupervised SimCSE.} adopts \textit{dropout} as data augmentation and encodes an input sentence \textit{twice} into two corresponding embeddings to build a positive pair. Since SimCSE is a Transformer-based encoder that directly encodes the length information of sentences through positional embeddings, the two embeddings in a positive pair contain the same length information.
Thus, a model trained with these positive pairs is biased, tending to consider that sentences of the same or similar length are more similar in semantics.
To alleviate it, we apply a simple but effective repetition operation to modify the input sentence. Then we pass the input sentence and its modified counterpart to the pre-trained Transformer encoder, respectively, to get the positive pair.
Additionally, we draw inspiration from the computer vision community and introduce momentum contrast to enlarge the number of negative pairs without additional calculations. 
The proposed modifications are applied to positive and negative pairs separately, and build a new sentence embedding method, termed Enhanced SimCSE (ESimCSE). 
We evaluate the proposed ESimCSE on several benchmark datasets w.r.t the semantic text similarity (STS) task. Experimental results show that ESimCSE outperforms SimCSE by an average Spearman correlation of 2.02\% on BERT-base. Our code are available at \href{https://github.com/caskcsg/ESimCSE}{https://github.com/caskcsg/ESimCSE}.
\end{abstract}

\section{Introduction}
% The large-scale pre-trained language model ~\cite{devlin2018bert,liu2019roberta},
% represented by BERT, benefits many downstream supervised tasks through finetuning methods. However, when applying BERT's native sentence embeddings directly for semantic similarity tasks \textit{without labeled data}, the performance is hardly satisfactory ~\cite{gao2021simcse,yan2021consert}. 
Recently, researchers have proposed using contrastive learning to learn better unsupervised sentence embeddings \cite{wu2020clear,zhang2020unsupervised,liu2021fast,gao2021simcse,yan2021consert}.
Contrastive learning aims to learn effective sentence embeddings based on the assumption that effective sentence embeddings should bring similar sentences closer while pushing away dissimilar ones. 
It generally uses various data augmentation methods \cite{shleifer2019low, wei2019eda, wu2019conditional} to generate different views for each sentence randomly, and assumes a sentence is semantically more similar to its augmented counterpart than any other sentence. 
Among these methods, the most representative one is SimCSE ~\cite{gao2021simcse}, which performs on par with previously supervised counterparts.
SimCSE implicitly hypothesizes \textit{dropout} acts as a minimal data augmentation method. 
Specifically,  SimCSE composes $N$ sentences in a batch and feeds each sentence to the pre-trained BERT \textit{twice} with two independently sampled dropout masks. 
Then the embeddings derived from the same sentence constitute a ``positive pair'', while those derived from two different sentences constitute a ``negative pair''. 

\begin{table}[!tbp]
\centering
\begin{tabular}{cl}
\midrule
\textbf{Length Diff}  & \textbf{Avg. Similarity Diff } \\
\midrule
 $$\textgreater$$ 3 & 16.34 \\
 $\leq$ 3 & \textbf{18.18} (+1.84) \\
\bottomrule
\end{tabular}
\caption{The average similarity difference between the model (SimCSE-BERT) predictions and the normalized ground truths.}
\label{cos_diff_simcse}
\end{table}

\begin{table*}[!htbp]
\centering
%\resizebox{.95\columnwidth}{!}{
\begin{tabular}{|l|p{10cm}|c|}
\hline
\textbf{Method}& \textbf{Text} & \textbf{Similarity}  \\
\hline
original sentence& I like this apple because it looks so fresh and it should be delicious. & 1.0  \\
\hline
random insertion& I \textbf{don't} like this apple because \textbf{but} it looks so \textbf{not} fresh and it should be \textbf{dog} delicious. & 0.69  \\
\hline
random deletion& I like this \sout{apple} because it looks so \sout{fresh} and it should be \sout{delicious}.  & 0.32  \\
\hline
word repetition& I like \textbf{like} this apple because it looks so \textbf{so} fresh and \textbf{and} it should be delicious. & 0.99  \\
\hline
word repetition& I \textbf{I} like this apple \textbf{apple} because it looks \textbf{looks} so fresh \textbf{fresh} and it should be delicious \textbf{delicious}. & 0.98  \\
\hline
\end{tabular}
\caption{An example of semantic similarity after different methods change a sentence's length. }
\label{table2}
\end{table*}

Using dropout as a minimal data augmentation method is simple and effective, but there is a weak point. 
SimCSE models are built on Transformer blocks, which will encode a sentence's length information through positional embeddings. In a positive pair, two embeddings are derived from the same sentence to contain the same length information. In contrast, in a negative pair, two embeddings in a negative pair are derived from two different sentences and generally contain different length information. Therefore, positive and negative pairs are different in their length information, acting as a feature to distinguish them. 
The semantic similarity model trained with these pairs can be biased, which probably considers that two sentences of the same or similar lengths are more similar in semantics.
To confirm it, we evaluate on seven standard semantic textual similarity datasets with the SimCSE-BERT$_{base}$ model published by ~\cite{gao2021simcse}. 
We partition each STS test set into two groups based on whether the sentence pairs' length difference is $\leq 3$.
We calculate the similarity differences between the model predictions and the normalized ground truths for each group. As shown in Table \ref{cos_diff_simcse}, the average similarity difference of seven datasets is higher when the length difference is $\leq 3$, which verifies our assumption. Comparison details on each dataset can refer to Table \ref{cos_diff_Esimcse2}.

\begin{figure*}
\centering
\includegraphics[width=12cm]{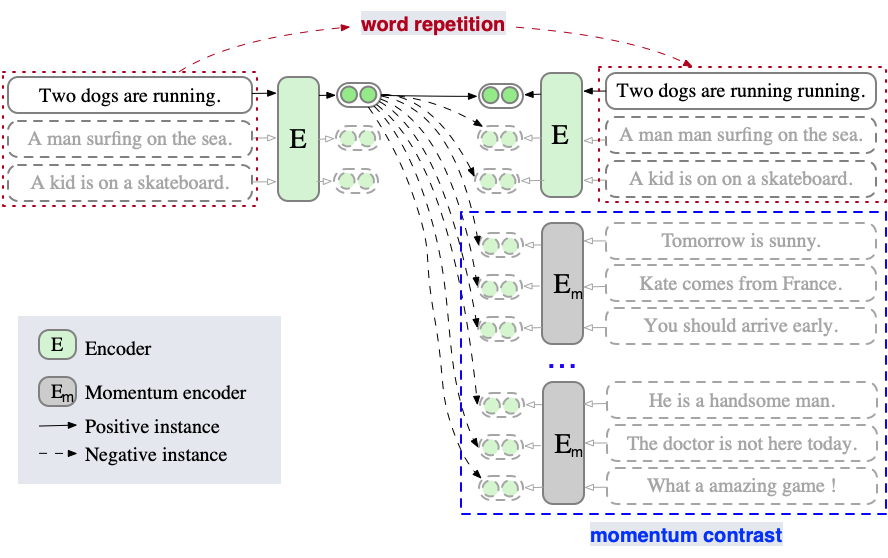}
\caption{The schematic diagram of the ESimCSE method.
% Unlike the SimCSE, ESimCSE performs word repetition operations on the batch so that the lengths of positive pairs vary without changing the semantics of sentences. This mechanism weakens the same-length hint for the model when predicting positive pairs. In addition, ESimCSE also maintains several preceding mini-batches' model outputs in a queue, termed momentum contrast, which can expand the negative pairs involved in loss calculation. This mechanism allows pairs to be compared more sufficiently in contrastive learning.
}
\label{fig_ESimCSE}
\end{figure*}

To alleviate this problem, we propose a simple but effective enhancement method to SimCSE. 
For each positive pair, we expect to change the length of a sentence without changing its semantic meaning. 
Existing methods to change the length of a sentence generally use random insertion and random deletion. 
However, inserting randomly selected words into a sentence may introduce extra noise, which will probably distort the meaning of the sentence; deleting keywords from a sentence will also change its semantics substantially.
Such operations are detrimental to SimCSE learning, which is also discussed in a contemporaneous work \cite{chuang2022diffcse}.
Therefore, we propose a safer method, termed ``word repetition'',  which randomly duplicates some words in a sentence. 
For example, as shown in Table 2, either random insertion or random deletion may generate a sentence that deviates far from the meaning of the original sentence. On the contrary, the method of ``word repetition'' maintains the meaning of the original sentence quite well.  

Apart from the optimization above for positive pairs construction, we further explore how to optimize the construction of negative pairs.
Since contrastive learning is carried out between positive pairs and negative pairs, theoretically, more negative pairs can lead to a better comparison between the pairs ~\cite{chen2020simple}.
And thus, a potential optimization direction is to leverage more negative pairs, encouraging the model towards more refined learning.
However, according to ~\cite{gao2021simcse}, larger batch size is not always a better choice. For example, for the SimCSE-BERT$_{base}$ model, the optimal batch size is 64, and other settings of the batch size will lower the performance. 
Therefore, we tend to figure out how to expand the negative pairs more effectively. 
In the community of computer vision, to alleviate the GPU memory limitation when expanding the batch size, a feasible way is to introduce the momentum contrast ~\cite{he2020momentum}, which is also applied to natural language understanding ~\cite{fang2020cert}.
Momentum contrast allows us to reuse the encoded embeddings from the immediate preceding mini-batches to expand the negative pairs by maintaining a queue. It always enqueues the sentence embeddings of the current mini-batches and meanwhile dequeues the ``oldest'' ones.
As the enqueued sentence embeddings come from the preceding mini-batches, we keep a momentum updated encoder by taking the moving average of its parameters and use the momentum encoder to generate enqueued sentence embeddings.
%In the SimCSE method, a sample will pass through the pre-trained BERT twice with different dropout masks to obtain two different but similar vector embeddings. 
%We want the enqueued sentence representation to be as good as possible, so we use the sentence representation derived without word repetition. 
Note that, we turn off \textit{dropout} when using the momentum encoder, which can narrow the gap between training and prediction.

The above two optimizations are proposed separately for building positive and negative pairs. We finally combine both with SimCSE, termed Enhanced SimCSE (ESimCSE). We illustrate the schematic diagram of ESimCSE in Figure \ref{fig_ESimCSE}. 
The proposed ESimCSE is evaluated on the semantic text similarity (STS) task with 7 STS-B test sets.
% Both word repetition and momentum contrast bring substantial improvements to SimCSE. 
Experimental results show that ESimCSE can improve the similarity measuring performance in different model settings over the previous state-of-the-art SimCSE. Specifically, ESimCSE gains an average increase of Spearman’s correlation over SimCSE by +2.02\% on BERT$_{base}$.

% \begin{figure}[!tbp]
% \centering
% \includegraphics[width=8cm]{batch_size.jpg}
% \caption{The performance trend on STS-B development set when batch size changes for SimCSE-BERT$_{base}$ model.}
% \label{fig_bs}
% \end{figure}

Our contributions can be summarized as follows:
\begin{itemize}
\item We observe that SimCSE constructs each positive pair with two sentences of the same length, which can bias the learning process. We propose a simple but effective ``word repetition'' method to alleviate the problem.
\item We propose to use the momentum contrast method to increase the number of negative pairs involved in the loss calculation, which encourages the model towards more refined learning.   
\item We conduct extensive experiments on several benchmark datasets w.r.t semantic text similarity task. The experimental results well demonstrate that both proposed optimizations bring improvements to SimCSE.
\end{itemize}

\section{Background: SimCSE}
Given a set of paired sentences $\left\{x_i, x_i^+\right\}_{i=1}^{m}$, where $x_i$ and $x^{+}_i$ are semantically related and will be referred to positive pairs. The core idea of SimCSE is to use identical sentences to build the positive pairs, i.e., $x^{+}_i = x_i$. Note that in Transformer, there is a dropout mask placed on fully-connected layers and attention probabilities. And thus, the key ingredient is to feed the same input $x_i$ to the encoder twice by applying different dropout masks $z_i$ and $z_i^{+}$ and output two separate sentence embeddings to build a positive pair as follows:
\begin{equation}
\mathbf{h}_{i}=f_{\theta}\left(x_{i}, z_i\right), 
\mathbf{h}_{i}^{+}=f_{\theta}\left(x_{i}, z_i^{+}\right)
\end{equation}
With $h_i$ and $h_i^{+}$ for each sentence in a mini-batch with batch size $N$, the contrastive learning objective w.r.t $x_i$ is formulated as follows,
\begin{equation}
\ell_{i}=-\log \frac{e^{\mathrm{sim}\left(\mathbf{h}_{i}, \mathbf{h}_{i}^{+}\right) / \tau}}{\sum_{j=1}^{N} e^{\mathrm{sim}\left(\mathbf{h}_{i}, \mathbf{h}_{j}^{+}\right) / \tau}}
\end{equation}
where $\tau$ is a temperature hyperparameter and $\mathrm{sim}\left(\mathbf{h}_{i}, \mathbf{h}_{i}^{\prime}\right)$ is  the  similarity metric,  which  is  typically the cosine similarity function.
% \begin{equation}
% \mathrm{sim}\left(\mathbf{h}_{i}, \mathbf{h}_{i}^{+}\right) = \frac{\mathbf{h}_{i}^{\top} \mathbf{h}_{i}^{+}}{\left\|\mathbf{h}_{i}\right\| \cdot\left\|\mathbf{h}_{i}^{+}\right\|}
% \end{equation}

% \begin{table*}[!htbp]
% \centering
% \begin{tabular}{cccccccc}
% \toprule
%  & \textbf{STS12} & \textbf{STS13} & \textbf{STS14} & \textbf{SICK15} & \textbf{STS16} & \textbf{STS-B} & \textbf{SICK-R} \\
%  \midrule
% train & 0 & 0 & 0 & 0 & 0 & 5,749 & 4,500 \\
%  \midrule
% dev & 0 & 0 & 0 & 0 & 0 & 1,500 & 500 \\
%  \midrule
% test & 3,108 & 1,500 & 3,750 & 3,000 & 1,186 & 1,379 & 4,927 \\
% \bottomrule  
% \end{tabular}
% \caption{Data statistics of standard semantic textual similarity (STS) tasks.}
% \label{data_statistics}
% \end{table*}

\section{Proposed Enhanced SimCSE}
In this section, we first introduce the word repetition method to construct better positive pairs. Then we introduce the momentum contrast method to expand negative pairs.
\subsection{Word Repetition}
The word repetition mechanism randomly duplicates some words/sub-words in a sentence. Here we take sub-word repetition as an example.
Given a sentence $s$, after processing by a sub-word tokenizer, we get a sub-word sequence $x = \{x_1, x_2, ..., x_N\}$, $N$ being the length of sequence. 
We define the number of repeated tokens as
\begin{equation}
dup\_len \in [0, max(2, int(dup\_rate * N))]
\end{equation}
where $dup\_rate$ is the maximal repetition rate, which is a hyperparameter. 
Then $dup\_len$ is a randomly sampled number in the set defined above, which will introduce more diversity when extending the sequence length. 
After $dup\_len$ is determined, we use uniform distribution to randomly select $dup\_len$ sub-words that need to be repeated from the sequence, which composes the $dup\_set$ as follows,
\begin{equation}
dup\_set = uniform([1,N], num=dup\_len)
\end{equation}
For example, if the $1$st sub-word is in $dup\_set$, then sequence $x$ becomes $x^{+} = \{x_1, x_1, x_2, ..., x_N\}$. And different from SimCSE which passes $x$ to the pre-trained BERT twice, E-SimCSE passes $x$ and $x^{+}$ independently.

\begin{table*}[!ht]
\centering
\setlength{\tabcolsep}{0.9mm}{\begin{tabular}{lcccccccl}
\toprule  
\textbf{Model}& \textbf{STS12} & \textbf{STS13} & \textbf{STS14} & \textbf{SICK15} & \textbf{STS16} & \textbf{STS-B} & \textbf{SICK-R} & \textbf{Avg.} \\
\midrule
\midrule
IS-BERT$_{base}$ $\triangle$ & 56.77 & 69.24 & 61.21 & 75.23 & 70.16 & 69.21 & 64.25 & 66.58 \\
CT-BERT$_{base}$ $\triangle$ & 61.63 & 76.80 & 68.47 & 77.50 & 76.48 & 74.31 & 69.19 & 72.05 \\
ConSERT$_{base}$ $\heartsuit$ & 64.64 & 78.49 & 69.07 & 79.72 & 75.95 & 73.97 & 67.31 & 72.74 \\
BERT$_{base}$-flow$\diamondsuit$ & 63.48 & 72.14 & 68.42 & 73.77 & 75.37 & 70.72 & 63.11 & 69.57 \\
SG-OPT-BERT$_{base}$ $\spadesuit$ & 66.84	& 80.13	& 71.23	& 81.56	& 77.17  & 77.23 & 68.16 & 74.62 \\
Mirror-BERT$_{base}$ $\sharp$ & 69.10 & 81.10 &  73.00 & 81.90 & 75.70 & 78.00 & 69.10 & 75.40 \\
SimCSE-BERT$_{base}$ $\clubsuit$ & 68.40 & 82.41 & 74.38 & 80.91 & 78.56 & 76.85 & 72.23 & 76.25 \\
ESimCSE-BERT$_{base}$ & \textbf{73.40} & \textbf{83.27} & \textbf{77.25} & \textbf{82.66} & \textbf{78.81} & \textbf{80.17} & \textbf{72.30} & \textbf{78.27}\\
\midrule 
ConSERT$_{large}$ $\heartsuit$ & 70.69 & 82.96 & 74.13 & 82.78 & 76.66 & 77.53 & 70.37 & 76.45 \\
BERT$_{large}$-flow$\diamondsuit$ & 65.20 & 73.39 & 69.42 & 74.92 & 77.63 & 72.26 & 62.50 & 70.76 \\
SG-OPT-BERT$_{large}$ $\spadesuit$ & 67.02 & 79.42 & 70.38 & 81.72 & 76.35 & 76.16 & 70.20 & 74.46  \\
SimCSE-BERT$_{large}$ $\clubsuit$ & 70.88 & 84.16 & 76.43 & \textbf{84.50} & \textbf{79.76} & 79.26 & 73.88 & 78.41 \\
ESimCSE-BERT$_{large}$ & \textbf{73.21} & \textbf{85.37} & \textbf{77.73} & 84.30 & 78.92 & \textbf{80.73} & \textbf{74.89} & \textbf{79.31} \\
\midrule 
Mirror-RoBERTa$_{base}$ $\sharp$ & 66.60 & 82.70 & 74.00 & 82.40 & 79.70 & 79.60 & 69.70 & 76.40 \\
SimCSE-RoBERTa$_{base}$ $\clubsuit$ & \textbf{70.16} & 81.77 & 73.24 & 81.36 & \textbf{80.65} &  80.22 & 68.56 & 76.57 \\
ESimCSE-RoBERTa$_{base}$ & 69.90 & \textbf{82.50} & \textbf{74.68} & \textbf{83.19} & 80.30 & \textbf{80.99} & \textbf{70.54} & \textbf{77.44}\\
\midrule 
SimCSE-RoBERTa$_{large}$ $\clubsuit$ & 72.86 & 83.99 & 75.62 & 84.77 & \textbf{81.80} & 81.98 &  71.26 & 78.90 \\
ESimCSE-RoBERTa$_{large}$ & \textbf{73.20} & \textbf{84.93} & \textbf{76.88} & \textbf{84.86} & 81.21 & \textbf{82.79} & \textbf{72.27} & \textbf{79.45} \\
\bottomrule
\end{tabular}}
\caption{Sentence embedding performance on 7 semantic textual similarity (STS) test sets. $\clubsuit$ : results from official published model by ~\cite{gao2021simcse}.$\heartsuit$ : results from ~\cite{yan2021consert}. $\spadesuit$ : results from ~\cite{kim2021self}. $\diamondsuit$ : results from ~\cite{li2020sentence}. $\triangle$ : results are reproduced and reevaluated by ~\cite{gao2021simcse}. $\sharp$ : results from ~\cite{liu2021fast}}.
\label{table_test_best}
\end{table*}

\subsection{Momentum Contrast}
The momentum contrast allows us to reuse the encoded sentence embeddings from the immediate preceding mini-batches by maintaining a queue of a fixed size. Specifically, the embeddings in the queue are progressively replaced. When the output sentence embeddings of the current mini-batch is enqueued, the ``oldest'' ones in the queue are removed if the queue is full. Note that we use a momentum-updated encoder to encode the enqueued sentence embeddings. Formally, denoting the parameters of the encoder as $\theta_{\mathrm{e}}$ and those of the momentum-updated encoder as $\theta_{\mathrm{m}}$, we update $\theta_{\mathrm{m}}$ in the following way,
\begin{equation}
\theta_{\mathrm{m}} \leftarrow \lambda \theta_{\mathrm{m}}+(1-\lambda) \theta_{\mathrm{e}}
\end{equation}
where $\lambda \in [0, 1)$ is a momentum coefficient parameter. Note that only the parameters $\theta_{\mathrm{e}}$ are updated by back-propagation. And here we introduce $\theta_m$ to generate sentence embeddings for the queue, because the momentum update can make $\theta_{\mathrm{m}}$ evolve more smoothly than
$\theta_{\mathrm{e}}$. As a result, though the embeddings in the queue are encoded by different encoders (in different ``steps'' during training), the difference among these encoders can be made small.

With sentence embeddings in the queue, the loss function of ESimCSE is further modifed as follows,
\begin{equation}
\ell_{i}=-\log \frac{e^{\mathrm{sim}\left(\mathbf{h}_{i}, \mathbf{h}_{i}^{+}\right) / \tau}}{\sum_{j=1}^{N} e^{\mathrm{sim}\left(\mathbf{h}_{i}, \mathbf{h}_{j}^{+}\right) / \tau}+ \sum_{m=1}^{M} e^{\mathrm{sim}\left(\mathbf{h}_{i}, \mathbf{h}_{m}^{+}\right) / \tau}}
\end{equation}
where $h_{m}^{+}$ is denotes a sentence embedding in the momentum-updated queue, and $M$ is the size of the queue.

\section{Experiment}
\subsection{Experiment Setup}
Our experimental language is English. For a fair comparison, our experimental setup mainly follows SimCSE. We use 1-million sentences randomly drawn from English Wikipedia for training\footnote{https://huggingface.co/datasets/princeton-nlp/datasets-for-simcse/resolve/main/wiki1m\_for\_simcse.txt}.
The semantic textual similarity task measures the capability of sentence embeddings, and we conduct our experiments on seven standard semantic textual similarity (STS) datasets. 
STS12-STS16 datasets  \cite{agirre2012semeval,agirre2013sem,agirre2014semeval,agirre2015semeval,agirre2016semeval}  do not have train or development sets, and thus we evaluate the models on the development set of STS-B \cite{cer2017semeval} to search for better settings of the hyper-parameters.
The SentEval toolkit\footnote{https://github.com/facebookresearch/SentEval} is used for evaluation, and Spearman correlation coefficient \footnote{\url{https://en.wikipedia.org/wiki/Spearman\%27s_rank_correlation_coefficient}}  is used to report the model performance. 
All the experiments are conducted on Nvidia 3090 GPUs. 

% For a set of size $n$, the n raw scores $X_{i}, Y_{i}$ are converted to its corresponding ranks $\mathrm{rg}_{X_{i}}, \mathrm{rg}_{Y_{i}}$, then the Spearman correlation is defined as follows 
% \begin{equation}
% r_{s}=\frac{\mathrm{cov}\left(\mathrm{rg}_{X}, \mathrm{rg}_{Y}\right)}{\sigma_{\mathrm{rg}_{X}} \sigma_{\mathrm{rg}_{Y}}}
% \end{equation}
% where $\mathrm{cov}\left(\mathrm{rg}_{X}, \mathrm{rg}_{Y}\right)$ is the covariance of the rank variables, $\sigma_{\mathrm{rg}_{X}}$ and $\sigma_{\mathrm{rg}_{Y}}$ are the standard deviations of the rank variables.
% Spearman correlation has a value between -1 and 1, which will be high when the ranks of predicted similarities and the ground-truth are similar.

\subsection{Training Details}
We start from pre-trained checkpoints of BERT(uncased) or RoBERTa(cased) using both the base and the large versions, and we add an MLP layer on top of the [CLS] representation to get the sentence embedding. We implement ESimCSE based on Huggingface's transformers package\footnote{https://github.com/huggingface/transformers,version 4.2.1.}. We train our models for one epoch using the Adam optimizer with the batch size $= 64$ and the temperature $\tau = 0.05$ in Eq. (3). The learning rate is set as 3e-5 for ESimCSE-BERT$_{base}$ model and 1e-5 for other models. 
The dropout rate is $p=0.1$ for base models, $p=0.15$ for large models.
For the momentum contrast, we empirically choose a relatively large momentum $\lambda$ = 0.995.
In addition, following SimCSE's code, we evaluate the model every 125 training steps on the development set of STS-B and keep the best checkpoint for the final evaluation on test sets. We use sub-word repetition instead of word repetition, further discussed in the ablation study section.

\subsection{Main Results}
% \begin{table}[!bp]
% \centering
% \begin{tabular}{ll}
% \toprule  
% \textbf{Model}& \textbf{STS-B} \\
% \midrule
% SimCSE-BERT$_{base}$$\clubsuit$ & 82.45\\
% ESimCSE-BERT$_{base}$ & \textbf{84.85} (+2.40) \\
% \midrule 
% SimCSE-BERT$_{large}$$\clubsuit$ & 84.41 \\
% ESimCSE-BERT$_{large}$ & \textbf{86.60} (+2.19) \\
% \midrule 
% SimCSE-RoBERTa$_{base}$$\clubsuit$ & 83.91 \\
% ESimCSE-RoBERTa$_{base}$ & \textbf{85.10} (+1.19) \\
% \midrule 
% SimCSE-RoBERTa$_{large}$$\clubsuit$ &  85.07\\
% ESimCSE-RoBERTa$_{large}$ & \textbf{85.33} (+0.26) \\
% \bottomrule  
% \end{tabular}
% \caption{Sentence embedding performance on semantic textual similarity (STS) development sets. $\clubsuit$ : results from official published model by \cite{gao2021simcse}.}
% \label{table_dev_best}
% \end{table}

% Table \ref{table_dev_best} shows the best results obtained on the STS-B development sets. We highlight the highest numbers among models with the same pre-trained encoder as bold. $\clubsuit$ denotes the evaluation results from the official published model by ~\cite{gao2021simcse}. It can be seen that our proposed ESimCSE outperforms SimCSE by +2.40\% on BERT$_{base}$, +2.19\% on BERT$_{large}$ , +1.19\% on RoBERTa$_{base}$ , +0.26\% on RoBERTa$_{large}$, respectively.

% The comparison between the proposed ESimCSE and SimCSE on the development set gives us the first glance at the superiority of the proposed ESimCSE. 

Table \ref{table_test_best} shows the models' performance on seven semantic textual similarity (STS) test sets. We mainly select SimCSE for comparison, since it is the current state-of-the-art and shares the same setting as our approach. In addition, we also use IS-BERT \cite{zhang2020unsupervised}, CT-BERT \cite{carlsson2021semantic}, ConSERT \cite{yan2021consert}, SG-OPT \cite{kim2021self}, BERT-flow \cite{li2020sentence}, Mirror-BERT \cite{liu2021fast} as baselines. It can be seen that ESimCSE improves the measurement of semantic textual similarity in different settings over SimCSE. Specifically, ESimCSE outperforms SimCSE by +2.02\% on BERT$_{base}$, +0.90\% on BERT$_{large}$ , +0.87\% on RoBERTa$_{base}$ , +0.55\% on RoBERTa$_{large}$, respectively.

% \begin{table*}[htbp]
% \centering
% \setlength{\tabcolsep}{0.9mm}{\begin{tabular}{lcccccccl}
% \toprule  
% \textbf{Model}& \textbf{STS12} & \textbf{STS13} & \textbf{STS14} & \textbf{SICK15} & \textbf{STS16} & \textbf{STS-B} & \textbf{SICK-R} & \textbf{Avg.} \\
% \midrule
% SimCSE-BERT$_{base}$ $\clubsuit$ & 68.40 & 82.41 & 74.38 & 80.91 & 78.56 & 76.85 & 72.23 & 76.25 \\
% + word repetition & 69.79 & \textbf{83.43} & 75.65 & 82.44 & 79.43 & 79.44 & 71.86 & 77.43 (+1.18) \\
% + momentum contrast & 71.41 & 82.23 & 74.94 & \textbf{82.99} & \textbf{79.85} & 79.48 & 71.85 & 77.54 (+1.29) \\
% ESimCSE-BERT$_{base}$ & \textbf{73.40} & 83.27 & \textbf{77.25} & 82.66 & 78.81 & \textbf{80.17} & \textbf{72.30} & \textbf{78.27} (+2.02) \\
% \bottomrule
% \end{tabular}}
% \caption{Improvements on 7 STS test sets that word repetition or momentum contrast brings to SimCSE.}
% \label{table_compare_test}
% \end{table*}

\begin{table}[!htbp]
\centering
\begin{tabular}{ll}
\toprule  
\textbf{Model}& \textbf{STS-B} \\
\midrule
SimCSE $\clubsuit$ & 82.45 \\
+ word repetition & 84.09 (+1.64)\\
+ momentum contrast & 83.98 (+1.53)\\
ESimCSE & \textbf{84.85} (+2.40)\\
\bottomrule
\end{tabular}
\caption{Improvement on STS-B development sets that word repetition or momentum contrast brings to SimCSE. $\clubsuit$ : results from official published model by ~\cite{gao2021simcse}.}
\label{table_compare_dev}
\end{table}

\section{Ablation Study}
This section investigates how different settings affect ESimCSE's performance. All results are compared on BERT$_{base}$ scale models and are evaluated on the development set of STS-B unless otherwise specified. 

\subsection{The Importance of Word Repetition and Momentum Contrast}
We explore how much improvement it can bring to SimCSE when only using word repetition or momentum contrast.
As shown in Table \ref{table_compare_dev}, either word repetition or momentum contrast can bring substantial improvements to SimCSE. It means that both proposed methods to enhance the positive pairs and negative pairs are effective. Better yet, these two modifications can be superimposed (ESimCSE) to get further improvements.

\begin{table}[!tbp]
\centering
\begin{tabular}{lc}
\toprule  
\textbf{Length-extension Method}& \textbf{STS-B} \\
\midrule
+Inserting Stop-words & 81.72 \\
+Inserting [MASK] & 83.08 \\
+Inserting Masked Prediction & 84.18 \\
+Word Repetition & 84.40 \\
+Sub-word Repetition & \textbf{84.85} \\
\bottomrule
\end{tabular}
\caption{Effects of sentence-length-extension method.}
\label{table_different_repetition}
\end{table}

\subsection{Effect of Sentence-Length-Extension Method}
In addition to sub-word repetition, we also explore three other methods to increase sentence length:
\begin{itemize}
\item \textbf{Word Repetition} is similar to sub-word repetition, except that the repetition operation occurs \textit{before} tokenization. For example, given a word ``microbiology'', word repetition will produce ``microbiology microbiology'', while sub-word repetition will produce ``micro micro \#\#biology'' or ``micro \#\#biology \#\#biology''.
\item \textbf{Inserting Stop-words} inserts a random stop-word after the selected word instead of repeating the selected word.
\item \textbf{Inserting [MASK]} inserts a [MASK] token after the selected word. We can regard [MASK] as a dynamic context-compatible word placeholder.
\item \textbf{Inserting Masked Prediction} inserts a [MASK] token after the selected word and uses the masked language model to predict the top-1 substitution. The substitution is used to replace the inserted [MASK] token. 
\end{itemize}

As shown in Table \ref{table_different_repetition}, sub-word repetition achieves the best performance, and word repetition can also bring a good improvement, which shows that more fine-grained repetition can better alleviate the bias brought by the length difference of positive pairs. Inserting [MASK] can also improve slightly, but inserting stop words will decrease the effect. Inserting masked prediction also brings a 
good improvement, but this method requires a pre-trained masked language model to predict replacements, bringing high additional computational overhead.

\subsection{Batching Sentences of Similar Length in Training}
Apart from sentence-length-extension methods, we explore whether batching sentences of similar length in training will alleviate the bias towards identical sequence length in inference. We divide training sentences into buckets by length and batch them within each bucket. We explore two different settings:
\begin{itemize}
\item We divide the training set into two coarse-grained buckets based on whether the sentence length is greater than $buc\_len$, where $buc\_len \in [3,8]$;
\item We divide the training set by sentence length into 6 fine-grained buckets: $\{\leq 3, 4, 5, 6, 7, \geq 8\}$, which we use $buc\_len = 3\sim8$ for short.
\end{itemize}
We list the experimental results in Table \ref{table_bucket_length}. Dividing the training set into buckets does not bring significant improvements and even decreases in some settings. We believe that after being divided into buckets, shuffle can only be performed within a bucket, leading to an insufficient comparison in contrastive learning. In contrast, the effect of word repetition is much better.

\begin{table}[!b]
\centering
\begin{tabular}{lcccc}
\toprule 
$buc\_len$ & wr & 3 & 4 & 5  \\
STS-B & 84.09 & 81.92 & 82.00 & 82.66 \\
\midrule
$buc\_len$ & 6  & 7 & 8 & $3\sim8$ \\
STS-B  & 82.00 & 82.13 & 83.00 & 82.18 \\
\bottomrule
\end{tabular}
\caption{Effects of different bucket lengths $buc\_len$. ``wr'' means using word repetition method instead of bucketing sentences. ``$3\sim8$'' means fine-grained buckets setting: $\{\leq 3, 4, 5, 6, 7, \geq 8\}$.}
\label{table_bucket_length}
\end{table}

\begin{table*}[!htbp]
\centering
\begin{tabular}{llllllllll}
\midrule
\textbf{Model} & \textbf{LD} & \textbf{STS12} & \textbf{STS13} & \textbf{STS14} & \textbf{SICK15} & \textbf{STS16} & \textbf{STS-B} & \textbf{SICK-R}  & \textbf{Avg.}  \\
\midrule
\multirow{2}{*}{SimCSE} & $$\textgreater$$ 3 & 8.93 & 15.74 & 11.90 & 19.68 & 28.91 & 21.33 & 7.86 & 16.34 \\
& $\leq$ 3 & 9.29 & 22.81 & 19.53 & 19.92 & 24.08 & 22.12 & 9.53 & 18.18 \\
\midrule
\multirow{2}{*}{ESimCSE} & $$\textgreater$$ 3 & 13.48 & 23.73 & 17.14 & 25.98 & 34.71 & 26.22 & 10.44 & 21.67 \\
 & $\leq$ 3 & 12.52 & 28.56 & 24.13 & 24.17 & 29.32 & 25.63 & 12.35 & 22.38 \\
\bottomrule
\end{tabular}
\caption{The difference between the model predicted cosine similarity and the true label on each dataset's test set. ``LD'' is short for length difference.}
\label{cos_diff_Esimcse2}
\end{table*}

\begin{table}[!htbp]
\centering
\begin{tabular}{lcc}
\toprule
\textbf{Model} & \textbf{ Sim \textless q,s1 $$\textgreater$$ } & \textbf{Sim \textless q,s2 \textgreater } \\
\midrule
SimCSE & 26.39 & 27.07\textbf{(+0.68)} \\
ESimCSE & 36.82  & 36.87\textbf{(+0.05)}      \\
\bottomrule
\end{tabular}
\caption{Effect of repeated words on the average similarity of two sets}
\label{tab:bias}
\end{table}

\subsection{The Relationship between The Similarity and Length Difference}
We further explore the relationship between the similarity and length difference of sentence pairs on ESimcSE, compared with that of SimCSE in the Introduction.
As STS12-STS16 datasets do not have train or development sets, and thus we evaluate the models on the test set of each dataset.
We partition each STS test set into two groups based on whether the sentence pairs' length difference is $\leq 3$. Then we calculate the similarity differences between the model predictions and the normalized ground truths for each group. As listed in Table \ref{cos_diff_Esimcse2}, ESimCSE significantly reduces the average similarity difference gap between \textgreater 3 and $\leq 3$, from 1.84 to 0.71, alleviating the learning bias we mentioned in the Introduction.

\subsection{Will Word Repetition Bring New Bias ? }
We further explore whether word repetition will mislead the model to be more inclined to consider sentences with repeated overlaps are more similar. We conduct a detection experiment on wiki data with the following settings:
\begin{enumerate}
  \item We randomly select a sentence as a query, such as q = ``I like \textbf{this} apple because it \textbf{looks} \textbf{very} fresh''
  \item  We use the query to randomly recall a candidate sentence with 13\%-17\% overlap tokens, such as s1 = ``\textbf{This} is a very tall tree and it \textbf{looks} \textbf{like} a giant''
  \item We apply the word-repetition operation on the overlap tokens in the candidate sentence and produce a word-repeated sentence, such as s2 = ``\textbf{This this} is a \textbf{very very} tall tree and it \textbf{looks looks} like a giant.''
  \item We calculate the similarity of \textless q, s1 \textgreater  and \textless q, s2 \textgreater and compare them.
\end{enumerate}
We experiment on 100 different query sentences and calculate their average similarity. As shown in Table \ref{tab:bias}, compared to the 0.68 increase of the SimCSE, ESimCSE-BERT only increased by 0.05. Therefore, word repetition does not bring a new bias to the learning process.

% \subsection{Effect of Dropout Rate}

% \begin{table}[!tbp]
% \centering
% \begin{tabular}{lccc}
% \toprule  
% $p$ & 0.1 & 0.15 & 0.2  \\
% \midrule
% STS-B & \textbf{84.85} & 84.75 & 83.37 \\
% \bottomrule
% \end{tabular}
% \caption{Effects of different dropout probabilities $p$ on the STS-B development set in terms of Spearman’s correlation.}
% \label{table_dropout_rate}
% \end{table}

% Dropout is the key ingredient to the SimCSE model, so different dropout rates $p$ are crucial to the model's performance. 
% According to ~\cite{gao2021simcse}, the optimal dropout rate for SimCSE-BERT$_{base}$ is $p = 0.1$. 
% Considering that ESimCSE additionally introduces word repetition and momentum contrast mechanisms, we re-examine the impact of different dropouts on its performance. 
% We experiment on three typical dropout rates, and the results are shown in Table \ref{table_dropout_rate}. 
% Specifically, when the dropout is 0.1, it achieves the best performance on the STS-B development set.
% When the dropout increases to 0.15, the performance is close to that of 0.1, with no significant drop.
% And even when the dropout reaches 0.2, the performance drops by nearly 1\%, but it still outperforms SimCSE.
% The experimental results kind of show the robustness of the superiority of the proposed ESimCSE over SimCSE, in terms of dropout rate.

\begin{table}[!tpb]
\centering
\begin{tabular}{lcccc}
\toprule 
$dup\_rate$ & 0.08 & 0.12 & 0.16  & 0.2 \\
STS-B & 83.5 & 83.62 & 82.01 & 83.01\\
\midrule
$dup\_rate$  & 0.24 & 0.28 & 0.32 & 0.36 \\
STS-B  & 84.24 & 82.96 & \textbf{84.85} & 83.84 \\
\bottomrule
\end{tabular}
\caption{Effects of repetition rate $dup\_rate$.}
\label{table_repetition_rate}
\end{table}

\subsection{Effect of Hyperparameters}
\paragraph{Repetition Rate} To quantitatively study the effect of repetition rate on the model performance, we slowly increase the repetition rate parameter $dup\_rate$ from 0.08 to 0.36, with each increase by 0.04. As shown in Table \ref{table_repetition_rate}, when $dup\_rate=0.32$, ESimCSE achieves the best performance, a larger or smaller $dup\_rate$ will cause performance degradation, which is consistent with our intuition. 
% Although there are small fluctuations, most of the results of the proposed ESimCSE still exceed the best results of SimCSE.

\paragraph{Momentum Queue Size} The size of the momentum contrast queue determines the number of negative pairs involved in the loss calculation. We experiment with the queue size equals to different multiples of the batch size. The experimental results are listed in Table \ref{table_m_c}. The optimal result is reached when the queue size was 2.5 times the batch size. A smaller queue size will reduce the effect. This is intuitive because more negative pairs participate in the loss calculation to compare positive pairs more fully. But a too large queue size also reduces the 
effect. We guess that is because the negative pairs in the momentum contrast are generated by the past ``steps'' during training, and a larger queue will use the outputs of more outdated encoder models which are quite different from the current one. And thus that will reduce the reliability of the loss calculation. 

\begin{table}[!t]
\centering
\begin{tabular}{lc}
\toprule  
\textbf{Queue Size}& \textbf{STS-B} \\
\midrule
$1  \times batch\_size$ & 83.83 \\
$1.5  \times batch\_size$ & 83.81 \\
$2  \times batch\_size$ & 83.03 \\
$2.5  \times batch\_size$ & \textbf{84.85} \\
$3  \times batch\_size$ & 82.66 \\
\bottomrule
\end{tabular}
\caption{Effects of queue size of momentum contrast.}
\label{table_m_c}
\end{table}

\begin{table*}[!htbp]
\centering
\begin{tabular}{lcccccccl}
\toprule 
\textbf{Model} & \textbf{MR} & \textbf{CR} & \textbf{SUBJ} & \textbf{MPQA} & \textbf{SST} & \textbf{TREC} & \textbf{MRPC} & \textbf{Avg.} \\
\hline 
  SimCSE $\clubsuit$ & 81.18 & \textbf{86.46} & 94.45 & \textbf{88.88} & 85.50 & 89.80 & 74.43 & 85.81 \\
 ESimCSE & \textbf{81.32} & 86.22 & \textbf{94.74} & 88.74 & \textbf{85.50} & \textbf{91.00} & \textbf{74.90} & \textbf{86.06} \\
\bottomrule
\end{tabular}
\caption{Results on transfer tasks of different sentence embedding models, in terms of accuracy. $\clubsuit$ : results from \cite{gao2021simcse}. 
}
\label{table_transfer}
\end{table*}

\subsection{Performance on Transfer Tasks}
Following \cite{gao2021simcse}, we further evaluate ESimCSE on transfer tasks, to see the transferability of the sentence embeddings from ESimCSE. 
The transfer tasks include: MR (movie review) \cite{pang2005seeing}, CR (product review) \cite{hu2004mining}, SUBJ (subjectivity status) \cite{ pang2004sentimental} , MPQA (opinion-polarity) \cite{wiebe2005annotating}, SST-2 (binary sentiment analysis) \cite{socher2013recursive}, TREC (question-type classification) \cite{voorhees2000building} and MRPC (paraphrase detection) \cite{dolan2005automatically}. For more details, one can refer to SentEval\footnote{\url{https://github.com/facebookresearch/SentEval} }. As shown in Table \ref{table_transfer}, compared with the performance of SimCSE, ESimCSE slightly increases the transferability of embedding. As our optimizations are focused on semantic textual similarity tasks, the ability of ESimcse on transfer tasks remains stable relative to SimCSE.

% \begin{table*}[htbp]
% \centering

% \begin{tabular}{llllllllll}
% \midrule
% \textbf{Model} & \textbf{length diff} & \textbf{STS12} & \textbf{STS13} & \textbf{STS14} & \textbf{SICK15} & \textbf{STS16} & \textbf{STS-B} & \textbf{SICK-R}  & \textbf{Avg.}  \\
% \midrule
% SimCSE-BERT$_{base}$ & \leq 3 & 70.62 & 83.76 & 77.99 & 82.61 & 79.16 & 78.50 & 69.90 & 77.51 \\
% ESimCSE-BERT$_{base}$ & \leq 3 & 76.16 & 84.63 & 80.76 & 84.42 & 80.18 & 82.49 & 70.28 & 79.85\textbf{(+2.34)}  \\

% \midrule
% SimCSE-BERT$_{base}$ & \textgreater 3 & 57.86 & 83.63 & 66.55 & 74.35 & 74.49 & 68.59 & 73.42 & 71.27  \\
% ESimCSE-BERT$_{base}$ & \textgreater 3 & 59.88 & 82.50 & 68.00 & 74.89 & 71.27 & 66.98 & 73.16 & 70.95\textbf{(-0.32)} \\

% \bottomrule
% \end{tabular}
% \caption{The spearman correlation of sentence pairs with a length difference of $\leq$ 3 and $\textgreater$ 3.}
% \label{spearman_diff}
% \end{table*}

\section{Related Work}
Unsupervised sentence representation learning has been widely studied. ~\cite{socher2011dynamic,hill2016learning,le2014distributed} propose to learn sentence representation according to the internal structure of each sentence. ~\cite{kiros2015skip,logeswaran2018efficient} predict the surrounding sentences of a given sentence based on the distribution hypothesis. ~\cite{pagliardini2017unsupervised} propose Sent2Vec, a simple unsupervised model allowing to compose sentence embeddings using word vectors along with n-gram embeddings. Recently, contrastive learning has been explored in unsupervised sentence representation learning and has become a promising trend \cite{zhang2020unsupervised,wu2020clear,meng2021coco,liu2021fast,gao2021simcse,yan2021consert,chuang2022diffcse}. Those contrastive learning based methods for sentence embeddings are generally based on the assumption that a good semantic representation should be able to bring similar sentences closer while pushing away dissimilar ones.
Therefore, those methods use various data augmentation methods to randomly generate two different views for each sentence and design an effective loss function to make them closer in the semantic representation space. Among these contrastive methods, the most related ones to our work are unsup-ConSERT \cite{yan2021consert} and unsup-SimSCE \cite{gao2021simcse}. ConSERT explores various effective data augmentation strategies(e.g., adversarial attack, token shuffling, Cutoff, dropout) to generate different views for contrastive learning and analyze their effects on unsupervised sentence representation transfer. Unsup-SimSCE, the current state-of-the-art unsupervised method uses only standard dropout as minimal data augmentation, and feed an identical sentence to a pretrained model twice with independently sampled dropout masks to generate two distinct sentence embeddings as a positive pair. Unsup-SimSCE is very simple but works surprisingly well, performing on par with previously supervised counterparts. 
However, we find that SimCSE constructs each positive pair with two sentences of the same length, which can mislead the learning of sentence embeddings. So we propose a simple but effective method temed ``word repetition'' to alleviate it. We also propose to use the momentum contrast method to increase the number of negative pairs involved in the loss calculation, which encourages the model towards more refined learning. 

\section{Conclusion and Future Work}
In this paper, we propose optimizations to construct positive and negative pairs for SimCSE and combine them with SimCSE, which is termed ESimCSE. Through extensive experiments, the proposed ESimCSE achieves considerable improvements on standard semantic text similarity tasks over SimCSE.

In the future, we will focus on designing a more refined objective function to improve the discrimination between different negative pairs. Also we will make attempt to optimize the performance on both semantic textual similarity tasks and transfer tasks.
\bibliography{anthology,custom}
\end{document}